# Reconstructing Maps from Text

**Johnathan E. Avery, Robert L. Goldstone, and Michael N. Jones**
Indiana University, Bloomington
[averjo][rgoldsto][jonesmn]@indiana.edu

**Abstract**

Previous research has demonstrated that Distributional Semantic Models (DSMs) are capable of reconstructing maps from news corpora (Louwerse & Zwaan, 2009) and novels (Louwerse & Benesh, 2012). The capacity for reproducing maps is surprising since DSMs notoriously lack perceptual grounding (De Vega et al., 2012). In this paper we investigate the statistical sources required in language to infer maps, and resulting constraints placed on mechanisms of semantic representation. Study 1 brings word co-occurrence under experimental control to demonstrate that direct co-occurrence in language is necessary for traditional DSMs to successfully reproduce maps. Study 2 presents an instance-based DSM that is capable of reconstructing maps independent of the frequency of co-occurrence of city names.

## Introduction

Distributional Semantic Models (DSMs) posit cognitive mechanisms to explain how humans construct semantic representations for words from statistical regularities in natural language. Typically, these models represent words as points in a high-dimensional vector space, and similarity between words is measured as the proximity in this semantic space. Latent Semantic Analysis (LSA; Landauer & Dumais, 1997) is the classic example of a DSM, but more modern versions span theoretically diverse learning mechanisms (see Jones et al., 2015 for a review). In general, DSMs have shown remarkable success at accounting for a broad range of semantic phenomena from relatively simple mechanisms.

One major criticism of DSMs is that their representations are amodal and are not grounded in perception or action (De Vega et al., 2012). Without perceptual grounding, DSMs may lack a necessary source of statistical information to fully represent semantic relationships between words. However, there is often strong alignment between the statistical distributions of words in a corpus and perceptual data (Riordan & Jones, 2011; Roads & Love, 2020).

A surprising early demonstration of the capacities of DSMs was presented by Louwerse & Zwaan (2009) where they reproduced the map of the USA by applying LSA to various large news corpora. Louwerse and Benesh (2012) followed up this study by demonstrating that even on the relatively small corpus of The Lord of the Rings trilogy, LSA was able to closely reproduce a map of Middle Earth. Contrary to the assumption that spatial representations are fundamentally perceptual and necessarily grounded in sensory modalities, the reproduction of maps demonstrates that spatial distributions are encoded in language, and that semantic processes are able to elicit these spatial distributions independent of perceptual grounding.

While recreating the map of Middle Earth is entertaining as a demonstration, it opens up more questions regarding the source of its success than it answers. Specifically, it is unclear what the statistical properties of the corpus are that enable external spatial distributions to be reconstructed from the text. It may be the case that the additional words that occur in the same context as the cities serve as additional dimensions along which the relationships between cities may be differentiated. Alternatively, cities which are near each other may be discussed in the same contexts more frequently. In this paper, we examine the hypothesis that cities that are near one another are discussed in the same context with greater frequency than cities that are separated by greater distances.

We generate artificial corpora describing randomly generated maps to compare whether DSMs are able to elicit spatial distributions independently of frequency of sampling. Specifically, we bring sampling under experimental control by manipulating whether a statement relating a pair of cities has a uniform probability of ending up in the final corpus. Uniform sampling is compared to distance-based sampling, where a statement relating a pair of cities has a higher probability of showing up in the final corpus based on their relative distance, such that nearby cities are likely to be in the corpus whereas distant cities are not. Study 1 found that no standard DSM was able to reproduce spatial locations independent of sampling.

The unexpected lack of success for DSMs to reproduce spatial distributions is likely because they represent an 'abstraction at learning' class of models (Jones, 2019) that mirrors the 'prototype-vs-exemplar' debate in the categorization literature. Many researchers have noted that the static word meanings produced by DSMs are ineffective at modeling the way word meanings change as a function of learning or retrieval processes. Hence, Study 2 uses a recently proposed 'abstraction at retrieval' DSM, based on the idea that semantic relationships are produced on the fly (Instance Theory of Semantics; ITS; Jamieson et al., 2018). Using ITS we demonstrate that a retrieval-based DSM is capable of accurately reconstructing spatial representations from corpora given uniformly sampled descriptions of cities.

## Methods

Our goal is to determine whether a DSM is able to reproduce a map from systematically constructed linguistic descriptions of the relationships between cities. Here, we briefly describe our steps for evaluating DSM performance, with a more in-depth description to follow. We start by generating a corpus of descriptions of a map. First, we randomly generate a set of maps of varying distributions of cities. Next, for every pair of cities, we generate statements that describe the relationship between the two cities. In particular, we use two sets of relationships: North, South, East, and West; and near and far. We sample the sets of descriptions either uniformly or based on distance in order to yield multiple corpora. Once we have the corpora, we test a set of DSMs for their ability to reproduce the original maps based on the linguistic descriptions of the relationships between cities. We follow an outline of the steps used in Louwerse and Zwaan (2009) and Louwerse and Benesh (2012) in order to move from a corpus of text to a two-dimensional map, with minor modifications to their process. In particular, we start by training a model on a given corpus. We then find the cosine similarity between the vectors representing each city, and convert the similarity into distance using Shepard's (1987) exponential law of generalization. We convert the distance matrix into a two-dimensional plot using multidimensional scaling. Finally, we compare the two-dimensional plot yielded by the DSM to the original map using bidimensional regression (Friedman &

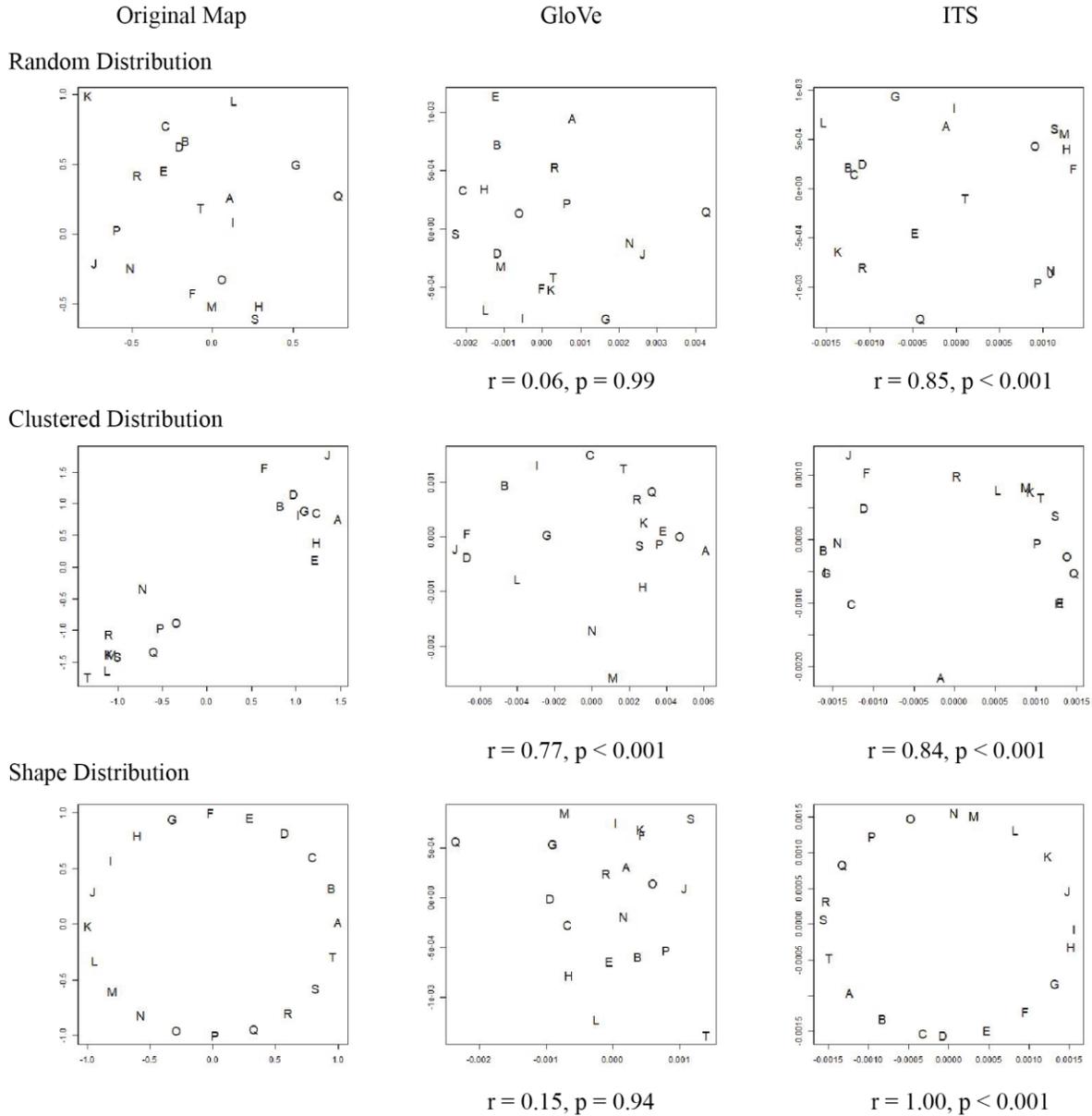

**Figure 1.** A representative sample of the performance of models at reconstructing the original map given the North/South/East/West relationship set with uniform sampling. From left to right, the columns present 1) the original maps, 2) the maps produced by GloVe (Study 1), and 3) the maps produced by ITS (Study 2). Each row shows one of the three distributions tested. The r coefficient and significance is noted below each map.

Kohler, 2003). Bidimensional regression is a special measure designed to compare two-dimensional plots that yields a measure of goodness of fit. Study 1 evaluates the performance of standard DSMs at reproducing maps given corpora where statements relating cities are sampled either uniformly or based on distance. Study 2 implements ITS coupled with a context retrieval mechanism, and evaluates this model's performance at reproducing maps as a function of sampling.

## Generate Maps

We generated three maps with different distributions of 'cities': random, clustered, and circular. The maps are displayed in the first column of Figure 1. There are 20 cities on each map, labelled with the letters 'A' through 'T'. We chose these three distributions of cities to vary the amount of external distribution information available to disambiguate the cities, and cover a range of challenges.

## Generate Corpora

We generated multiple sets of corpora for each combination of conditions. The conditions include the distribution of cities (random, clustered, shape), the description set (either North/South/East/West, or near/far), and the type of sampling (uniform or distance-based). In total, there were 12 corpora.

In order to generate the corpora, we created statements about each pair of cities on a given map. The statements took the form: [city] [relationship to] [city]. We used two sets of relationships: near/far, and North/South/East/West.

**Near/Far.** For each map, we found the average distance between cities. Subsequently, when generating a statement relating two cities (e.g. cities 'A' and 'B'), if the distance between the two cities was above the average distance we would label the relationship between the two cities as 'far' (e.g. 'A is far from B'). Conversely, if the distance between the two cities was lower than the average distance between all cities, we would label the relationship between the two cities as 'near' (e.g. 'A is near to B').

**North/South/East/West.** In order to formalize the directional relationship between cities, we divided the area surrounding a city into four quadrants such that a quadrant was 90 degrees, offset by 45 degrees. If some city 'B' appeared in the quadrant above a city 'A', it would be labeled as 'north of' city 'A' (e.g. 'B is North of A').

**Generating the statements.** We compared each city to every other city in order to generate a sentence regarding the two cities' relationships. Therefore, there are two sentences relating any two cities – with the relationship reversed. If the relationship between cities A and B is that 'A is North of B', then there will necessarily be a converse statement in the corpus that 'B is South of A'. There are no statements relating a city to itself (e.g. 'A is South of A').

A total of 20*19 = 380 possible unique statements relating all the cities to one another were generated. In the statements, the phrases were condensed such that the relationship was just one word (e.g. 'is North of' was reduced to 'north_of').

**Sampling the statements.** A corpus was generated by sampling the statements either uniformly or based on distance. Each corpus totaled 10,000 statements. If the set of statements was sampled uniformly, then each statement occurred an equal number of times in the final corpus. If the set of statements was sampled based on distance, statements relating cities that were closer together were sampled more frequently than statements relating cities that were farther. The probability of sampling a statement $t$ based on distance between two cities $i$ and $j$ is given by

$$P(t_{ij}) = \frac{d_{ij}}{\Sigma_i^I \Sigma_j^J d_{ij}} \quad (1)$$

where $d_{ij}$ is the distance between cities $i$ and $j$.

## Vector Spaces Produced by DSMs

We trained a set of models on each corpus. The set of models includes LSA (Landauer et al., 1998), Positive Pointwise Mutual Information (PPMI; Bullinaria & Levy, 2007), Continuous Bag Of Words (CBOW; Mikolov et al., 2013), and Global Vectors for word representation (GloVe; Pennington et al., 2014) – to be described in further detail in Study 1 – and the exemplar model, ITS (Jamieson et al., 2018) – to be described in further detail in Study 2.

Each of these models – with the exception of PPMI – operates by building a vector space, where words are placed in an arbitrary high-dimensional space. Their location in that space determines their similarity to the other words in that space, such that words that are near to each other are similar, while distant words are dissimilar. The cosine between two vectors is a commonly used metric to compute word similarities (Bullinaria & Levy, 2007).

Rather than working with similarities, the steps that follow require distances. We transform the cosine similarity matrix into a distance matrix using Shepard's (1987) universal law of generalization, such that the distance $d$ between two word vectors $i$ and $j$ is given by

$$d_{ij} = e^{-\gamma \, sim(i,j)}$$

where $sim(i, j)$ is the similarity between two word vectors $i$ and $j$, and $\gamma$ is a monotonic scaling factor.

The cosine similarities were computed between each city, and the similarity matrix was transformed into a distance matrix. Our processes yielded a distance matrix for each model, for each sampling method, for each description set, for each map, for a total of 60 distances matrices for evaluation.

## Generating and Evaluating the Map

We used multidimensional scaling (MDS) to generate a 2-dimensional map from the distance matrix. We then used bidimensional regression to evaluate the map produced by the model against the original map.

**MDS.** MDS is a well-established technique to transform a distance matrix into a plot of some arbitrary dimensionality (Kruskal & Wish, 1978). We used Kruskal's non-metric MDS as implemented in R (MASS; Venables & Ripley,

**Table 1.** Presents the r coefficients for each model at reproducing a map distribution given a particular sampling procedure and relationship set. The Near/Far relationship set is denoted 'N/F', while the North/South/East/West relationship set is denoted 'N/S/E/W'.

|  |  | Uniform | | Distance | |
| --- | --- | --- | --- | --- | --- |
| **Model** | **Map Distribution** | **N/F** | **N/S/E/W** | **N/F** | **N/S/E/W** |
| LSA | Random | 0.23 | 0.20 | 0.87*** | 0.86*** |
|  | Clustered | 0.41 | 0.25 | 0.69*** | 0.76*** |
|  | Circular | 0.34 | 0.21 | 0.09 | 0.76*** |
| CBOW | Random | 0.21 | 0.17 | 0.67*** | 0.72*** |
|  | Clustered | 0.11 | 0.77*** | 0.97*** | 0.85*** |
|  | Circular | 0.45 | 0.23 | 0.99*** | 0.76*** |
| GloVe | Random | 0.13 | 0.06 | 0.88*** | 0.88** |
|  | Clustered | 0.36 | 0.77*** | 0.97*** | 0.97*** |
|  | Cicular | 0.37 | 0.15 | 1.00*** | 1.00*** |
| PPMI | Random | 0.40 | 0.11 | 0.96*** | 0.96*** |
|  | Clustered | 0.29 | 0.12 | 0.97*** | 0.97*** |
|  | Circular | 0.47 | 0.43 | 1.00*** | 1.00*** |

p < 0.05 - *; p < 0.01 - **; p < 0.001 - ***

2002). Non-metric MDS produces a solution that matches the ordinal ranking of the distances provided in the distance matrix. The solutions produced by non-metric MDS are subject to rotations, shifts, scaling, and flips.

**Bidimensional Regression.** We used bidimensional regression to evaluate how well each MDS solution recreated the map from which it was derived (Friedman & Kohler, 2003; Louwerse & Zwaan, 2009). Bidimensional Regression is a measure of how well two maps align. The measure of interest for this study is the r coefficient. The r coefficient indicates how well two spatial distributions of points match each other. We generated our r coefficient using the affine bidimensional regression in R (BiDimRegression; Carbon, 2013).

## Study 1 – Standard DSMs

The purpose of Study 1 is to establish the source of success for DSMs in producing maps as demonstrated by Louwerse and Zwaan (2009) and Louwerse and Benesh (2012). We aimed to test the hypothesis that the first order co-occurrence of words drives the performance of modern DSMs. We manipulated the frequency by which statements are sampled.

### DSMs for evaluation

We first evaluate the ability of four state-of-the-art DSMs to produce spatial distributions from linguistic descriptions of maps[1]. While the set of models used here is not exhaustive, it is representative of the range of techniques of the best-performing models used in the field of semantic memory.

**LSA.** LSA is a well-established method of creating a semantic space by applying a combination of tf-idf and singular value decomposition (SVD) to a word-by-document frequency count of words in a text corpus (Landauer et al., 1998). SVD is a linear algebra technique that smooths a given matrix on the basis of its eigenvalues, or principle sources of variance. We used the python package, gensim (Řehůřek & Sojka, 2010), to implement LSA.

**CBOW.** CBOW is a neural network implementation of a DSM (Mikolov et al., 2013), where the localist input layer is a given word, and the output layer is the set of words in whose context the input word is found, with a single hidden layer. The algorithm finds the appropriate weights between nodes by minimizing the error between the context that the model predicts and the actual context in which a word occurs. The weights between nodes are treated as word embeddings in a high dimensional vector space. We used the CBOW implementation provided by the Python gensim package, with 50 training iterations, and the standard 300 nodes in the hidden layer.

**GloVe.** GloVe generates word vector representations using a gradient descent technique (Pennington et al., 2014). The technique minimizes the spatial distributions between the co-occurrence matrix and the vector space of arbitrarily high dimensionality. We used the GloVe implementation provided by Pennington et al., with 50 dimensions and a maximum 15 iterations to convergence.

**PPMI.** PMI is a log transform of the conditional probabilities of the co-occurrence of words (Bullinaria & Levy, 2007). Positive PMI sets negative PMI values to zero. PPMI is distinct from the other DSMs, as it produces a value relating two words based on their conditional probability, skipping the step in vector space models in which a word is placed in a high dimensional space and the relationship between words is derived from their cosine similarity. We used the PPMI implementation provided in the Python gensim package.

### Results

Table 1 presents the performance of each model at reproducing the original map given the different corpus sampling conditions for each relationship set. The significance of how well the map is reconstructed is indicated with asterisks. When the corpus is generated using uniform sampling, only a few models produced maps that significantly reproduced the original map. In contrast, when a corpus is generated with distance-based sampling, all models produce significant reproductions of the maps.

---
[1] Code for the analyses presented here available at: https://github.com/masterccioli/reconstructing-space-from-text

The center column in Figure 1 provides a visual demonstration of the performance of GloVe at reproducing the maps given the North/South/East/West relationship set with uniform sampling. Notably, the reproduction of the two clusters by GloVe demonstrates that though the reproduction is significant, a value of r = 0.77 corresponds to only a rough approximation of the original space. Note how in this example, the two clusters are not well separated. Overall, models using corpora generated by uniform sampling are mediocre at best at reconstructing the maps.

### Discussion

The manipulation of the frequency by which statements are sampled illuminates the source of success demonstrated by Louwerse and Zwaan (2009) and in Louwerse and Benesh (2012). Namely, DSMs require frequency of co-occurrence to encode geographical relations. Generally, objects that are highly related to one another are more likely to be discussed in the same context. This principle extends to the spatial locations of cities, such that cities that are spatially co-located are discussed in the same context more frequently. When we control for the frequency of co-occurrence, modern DSMs are not able to accurately co-locate cities in semantic space.

## Study 2 – A Retrieval-Based DSM

Study 1 demonstrates that standard DSMs depend on frequency of co-occurrence to reconstruct accurate spatial relationships. Specifically, modern DSMs treat frequently co-occurring words as more similar to one another.

Study 2 starts from the premise that humans may create semantic abstractions as a by-product of an episodic retrieval mechanism, and that word meanings depend on the context of retrieval. We use ITS (Jamieson et al., 2018) to demonstrate that a model with a sufficiently complex learning, storage, and retrieval process is capable of learning the spatial distributions of the three maps from linguistic descriptions independent of sampling procedure in corpus generation.

### Model

ITS uses a multiple-trace episodic memory store. Each instance is a set of words that co-occur in the same context, and is uniquely stored as a trace in memory. Memory may then be probed in order to get the vector representation of the word. When memory is probed, traces that contain the probe are recalled and combined into an echo. The echo is the normalized sum of all the contexts in which the probe word occurred (cf. Hintzman, 1986). When the probe is composed of multiple words, the echo is composed only of contexts where both words occur.

Table 2. Presents the r coefficients for ITS at reproducing a map distribution given a particular sampling procedure and relationship set.

| Relationship | Distribution | Uniform | Distance |
|---|---|---|---|
| Near/Far | Random | 0.83*** | 0.84*** |
|  | Clustered | 0.97*** | 0.97*** |
|  | Shape | 1.00*** | 1.00*** |
| North/S/E/W | Random | 0.85*** | 0.98*** |
|  | Clustered | 0.84*** | 0.85*** |
|  | Shape | 1.00*** | 0.99*** |

p < 0.05 - *; p < 0.01 - **; p < 0.001 - ***

An advantage of using an abstraction-at-retrieval model lies in the flexibility of the retrieval process. Specifically, ITS can take context of retrieval into account, a key drawback of DSMs pointed out by numerous researchers (Jamieson et al., 2018; Jones, 2019; Kintsch, 2000). We modify the retrieval process in two ways in order to maximize ITS ability to vary word meanings based on context. First, we modify the echo to yield simply the context of the probe without the probe itself. Second, we modify the process by which similarity is evaluated.

**Context of Probe.** In ITS, the retrieved echo is composed of the vectors representing the set of words that co-occur with the probe, as well as the probe itself. Here, we are interested in comparing the context in which a word occurs. Since the echo contains both the probe and the context of the probe, we must separate the context from the probe. Specifically, we define the context of a probe as the echo of a probe without the probe itself. Formally,

$$cont(probe) = echo(probe) - probe \qquad (2)$$

where *cont* denotes the context, and *echo(probe)* yields the echo given a probe as defined by ITS.

**Retrieval Process.** Ubiquitously, the similarity between words is treated as a direct cosine comparison between the vectors representing two words. Here, we deviate from convention by treating similarity as the comparison between the two words and some set of tertiary words. That is, the similarity between two words A and B can be approximated as what is shared between the context of word A and some other word C and the context of word B and that same word C. Words A and B are similar to the extent that the features that constitute the context of words A and C are shared with the features that constitute the context of words B and C. This definition of similarity does not deviate from the distributional hypothesis, but rather serves as an alternate formalization.

We define the similarity *sim* between two words *a* and *b* as

$$sim(a,b) = \sum_i^I cosine(cont(a,i), cont(b,i)) \qquad (3)$$

where *I* is the set of all unique words used in the corpus, *cosine* is the cosine similarity.

Here we present a concrete example of how such a retrieval process in ITS might elicit a spatial distribution of cities from a linguistic description of their relationships. Consider three cities A, B, and C, such that A and B are near each other and are both far from C. When comparing A and B, we use all the words in the corpus except A and B. For instance, we want to compare the *cont(A, C)* with *cont(B, C)*. The *cont(A, C)* yields 'far_from', likewise the *cont(B, C)* also yields 'far_from'. Therefore, when evaluating *cosine(cont(A, C), cont(B, C))*, the yielded value is high because the context for both pairs is identical.

In contrast, if we want to compare cities A and C, we would use B as a tertiary word of comparison. In this case, *cosine(cont(A, B), cont(C, B))* would yield a low similarity because the *cont(A, B)* yields 'near_to' while the *cont(C, B)* yields 'far_from'. Thusly, the process of eliciting the similarity between two words via the shared contexts with tertiary words can accurately elicit spatial locations from linguistic descriptions.

### Results

Table 2 presents the performance of ITS at reproducing the original map given the different corpus sampling conditions for each relationship set. The significance of how well the map is reconstructed is indicate with asterisks. Independent of sampling condition, ITS is able to produce significant reconstruction of all original maps. Notably, the lowest performing reconstruction has an r = 0.80, well above the highest performing model in Study 1 in the uniform sampling condition.

The right column in Figure 1 provides a visual demonstration of the performance of ITS at reproducing the maps given the North/South/East/West relationship set with uniform sampling. The ITS reconstruction of the shape provides compelling visual evidence that DSMs are capable of reproducing maps given different model assumptions.

## General Discussion

Demonstrations by Louwerse and Zwaan (2009) and Louwerse and Benesh (2012) show that spatial distributions can be elicited from text. Given that amodal DSMs are not grounded by any perceptual input, it is surprising that DSMs can reproduce spatial distributions whatsoever. Here we bring sampling frequency under experimental control to demonstrate that frequency of co-occurrence provides the statistical redundancy that enable standard DSMs to reproduce spatial distributions.

Study 1 demonstrates that standard "absraction-at-learning" DSMs are only able to elicit spatial distributions from linguistic descriptions given appropriate frequency of sampling. When cities that are near each other are discussed more frequently than cities that are far from each other, modern DSMs are able to reproduce their spatial distributions. When cities are discussed with uniform frequency, standard DSMs are not able to reproduce spatial distributions.

Study 2 explores a cognitively inspired 'abstraction-at-retrieval' DSM, where the semantic relationship between two words is dependent on the context in which the words co-occur, unlike 'abstraction-at-learning' DSMs. In an instance-based DSM, there is no stored semantic memory, only episodic memory. Semantic representations are constructed on-the-fly as an artifact of the episodic retrieval mechanism in response to an environmental probe. The model presented in Study 2 demonstrates that an instance-based DSM is capable of reproducing spatial distributions given uniformly sampled descriptions of cities.

Both Studies 1 and 2 demonstrate that spatial information can be elicited by DSMs independent of any grounding in the external world. While modern DSMs are capable of producing spatial distributions with distance-based sampling, it is insufficient to posit that humans are only capable of producing spatial distributions given linguistic descriptions based on frequency alone. Study 2 extends the representation of space toward a comprehension process, whereby spatial distributions can be elicited by DSMs independent of the first order co-occurrence of cities.